\documentclass[10pt]{article}
\usepackage[preprint]{acl}
\usepackage[T1]{fontenc}
\usepackage[utf8]{inputenc}
\usepackage{times}
\usepackage{microtype}
\usepackage{booktabs}
\usepackage{graphicx}
\usepackage{tabularx}
\usepackage{array}
\usepackage{dcolumn}
\usepackage{multirow}
\usepackage{amsmath}
\usepackage{xcolor}
\usepackage{enumitem}
\usepackage{cleveref}
\usepackage{orcidlink}
\hypersetup{hidelinks}
\setlist{nosep}
\hypersetup{
  pdftitle={Selection Is Retrieval, Abstention Is Not: On-Device Tool Routing over 70 Korean-English Actions},
  pdfauthor={Janghoon Lee},
  pdfsubject={On-device LLM routing, tool selection, abstention},
  pdfkeywords={on-device LLM routing, tool routing, abstention, BM25, multilingual retrieval}}
\newcommand{\code}[1]{\texttt{#1}}

\newcolumntype{Y}{>{\raggedright\arraybackslash}X}
\newcolumntype{S}[1]{D{.}{.}{#1}}
\crefname{section}{Section}{Sections}
\crefname{table}{Table}{Tables}
\crefname{figure}{Figure}{Figures}

\title{Selection Is Retrieval, Abstention Is Not:\\
On-Device Tool Routing over 70 Korean-English Actions}
\author{Janghoon Lee\,\orcidlink{0009-0002-8108-5407} \\
Redrob \\
\texttt{janghoon@redrob.io} \\
{\small ORCID 0009-0002-8108-5407}}
\date{}

\begin{document}
\maketitle

\begin{abstract}
An AI assistant that calls tools makes two decisions on every request: which
tool to invoke, and whether any available tool applies. In the usual design a
single language model makes both, by emitting a call or by declining to emit
one. On a device that has to answer without a server, the language model is
what makes that design expensive, dominating both the latency and the memory
of the router. The common alternative is to remove the model completely and
rank the catalog of local actions with a retriever instead.

That substitution is not symmetric across the two decisions. A retriever
returns its highest-scoring candidate for every input and cannot signal that
the catalog holds no valid action. Our earlier study found that constraining a
decoder to a tool grammar repairs malformed output without improving the
choice. What the substitution costs in each decision has not been measured.

We evaluate the two decisions separately over 600 Korean and English requests
and a catalog of 70 local actions. The router may also ask for a missing slot,
reply, or delegate. Half the in-catalog requests reuse catalog vocabulary and
half paraphrase it, separating lexical overlap from the action requested.

Character 3-gram BM25 selects 162 of 164 lexically matched requests and 85 of
166 paraphrases. Restricting the candidate set to seven raises the paraphrase
figure to a mean of 0.825 over five trials. No classifier over its score
features separates in-catalog from out-of-catalog above 0.697 area under the
curve, where the frozen encoder multilingual-e5-base reaches 0.806. Using that
encoder for abstention alone keeps 376 of the requests local and misroutes 9
of the 150 needing delegation. Abstention, not selection, is where a neural
component is required. A neural ranker improves every quality metric and is
rejected on latency and memory rather than accuracy.
\end{abstract}

\section{Introduction}
\label{sec:intro}

An AI assistant that calls tools makes two decisions on every request: which
tool to invoke, and whether any available tool applies. In the usual design a
single language model makes both, by emitting a call or by declining to emit
one. On a device that has to answer without a server, the language model is
what makes that design expensive. The catalog index used here occupies 17 KB
and returns a ranking in 0.140 ms (Section~\ref{sec:selection}), against
1,111 ms for a 4-billion-parameter caller on a GPU (Section~\ref{sec:deploy}).
The common alternative is therefore to remove the model completely and rank the
catalog of local actions with a retriever instead. The router still produces
one of four outcomes: a local action, a request for a missing slot, a
conversational reply, or delegation to a server.

That substitution is not symmetric across the two decisions, and the asymmetry
is what this paper measures. Selection asks which local action is nearest to
the request. Abstention asks whether the catalog contains a valid action at
all. A retriever answers the first by construction and cannot answer the
second, because it returns a nearest item even when every item is wrong.
Treating the two as a single multiclass problem conceals that asymmetry. We
evaluate each decision separately and report what the non-neural substitute
costs in each.

The first explanation to eliminate is malformed output, and it is not the
cause. Our earlier study constrained the decoder to the tool grammar and found
that it repaired output form while leaving accuracy among schema-compliant
calls unchanged \citep{lee2026repair}. The remaining error lies outside
formatting, and this paper locates it in the router itself, on a machine small
enough that the router cannot be a large model.

Component-level scoring is by now the standard evaluation practice, and it is
the practice this paper follows. \citet{bhat2026validity} separate tool
invocation, completion and outcome verification. FunctionChat-Bench distinguishes tool call, answer completion,
slot question and relevance detection in Korean dialogs
\citep{lee2024functionchat}. P-C-G separates planning, calling and generation
for Korean tool use, where 8B multi-chain call accuracy falls to 33.8\% from
95.6\% single-chain \citep{jeon2025pcg}. Two systems already split the two
decisions this paper is about. TinyAgent turns tool retrieval into
classification before generation \citep{erdogan2024tinyagent}, and a shipped
45M tool-calling model gives selection and abstention separate heads, one for
retrieval and one for calibrated confidence \citep{needle2026}. Both build the
split into the architecture. Neither reports what is lost when a compact
non-neural component takes one of the two roles, which is the measurement this
paper makes over a fixed catalog. Our companion study of server-side model
routing finds the same shape one layer up, where task type accounts for most of
the routable gap and a static task-by-language table captures it without a
learned router \citep{lee2026routing}.

The experiments establish four results.
\begin{enumerate}
  \item Separating an in-catalog gate from nearest-item selection reduces
  delegate-to-skill errors from 134 to 12 of 150
  (\cref{sec:different,tab:pool}).
  \item Lexical retrieval resolves selection under matched vocabulary, but
  BM25 score features do not resolve abstention
  (\cref{sec:selection,sec:abstention}).
  \item A neural ranker resolves nine more requests than the retained pipeline
  on accuracy and eight more on completion, and is rejected at 14 times that
  pipeline's CPU median and three times its memory. Encoder reranking recovers
  less of the shortlist advantage than native tool calling, and neither replaces
  the split architecture (\cref{sec:rerank}).
  \item Vocabulary match and active candidate count determine the measured
  ranker ordering, while two of four native-calling checkpoints do not deploy
  under the fixed protocol (\cref{sec:deploy,sec:choice}).
\end{enumerate}

\section{Setup}
\label{sec:setup}

\paragraph{Action space.}
The action space separates 70 searchable local actions from three outcomes.
The local actions contain 64 skills and six file tools. The three labels are
\code{chat}, \code{ask}, and \code{delegate}. The search pool contains the 70
skills and file tools. The gate decides in versus out before a fixed rule
separates chat from delegation. A local in decision invokes the selected
action or asks for a required slot.
We use \emph{abstention} for this in-versus-out decision. Delegation is one
possible final action after the router abstains.

Catalog text and execution contracts remain separate throughout the study.
A retrieval document contains a name, description, and optional aliases in
Korean and English. A contract contains typed slots, required flags, input and
output types, and mutation status. We call the base document index E0 and the
alias-expanded index E1. E1 changes only retrieval text. It never changes the
callable contract.
Appendix~\ref{app:tools} lists the nine non-skill actions and
Appendix~\ref{app:catalog} reports catalog length.

\paragraph{Utterances.}
The evaluation isolates vocabulary match from requested action. We authored
300 Korean and English pairs, for 600 requests total. The construction is
deliberate and is not a sample of production traffic. Of 330 skill or file-tool
utterances, 164 use lexical match and 166 use lexical mismatch. Match items use
words close to the catalog document. Mismatch items preserve the intended
action while changing the surface vocabulary. The remaining requests contain
60 ask, 60 chat, and 150 delegate labels. Ask is in-catalog because the local
action is known and only a required slot is missing. The out set therefore
contains 210 requests: 60 chat and 150 delegate. We call match items lexically
matched and mismatch items paraphrases throughout.

\paragraph{Verification.}
Deterministic post-state checks define execution success. Gold action IDs are
assigned in the authored ledger. We do not use an LLM judge. This choice avoids
judge variance. Deterministic evaluators can still fail when their state
predicates are brittle
\citep{bhat2026validity}. We inspect the contracts and post-state conditions
rather than comparing free-form text.

\paragraph{Metrics.}
Four metrics, two enforced caps, and one preregistered coverage target expose
both excessive delegation and unsupported execution.
Accuracy is exact final-action accuracy. Completion counts an exact local
completion and also a safe delegation for an executable local task. Local
coverage is $1-\Pr(\mathrm{delegate})$. Danger rate is the fraction of the 150
gold-delegate utterances sent to any skill or file tool. We constrain danger
to at most 9 of 150 and the ask rate to at most 120 of 600. The harness enforces
those two caps and then maximizes coverage. The preregistered target of 0.628
was not enforced, and the retained pipeline sits one request below it
(Limitations). These metrics prevent a router from
improving apparent accuracy by delegating everything, or coverage through
unsupported execution.

\paragraph{Cost axes.}
The preregistration named CPU latency, parameter bytes, and model count as
comparison axes without fixing a limit on any of them, so we report all three
for every pipeline and state explicitly where a quality gain is rejected on cost
rather than on a metric.
The routing decision has to complete inside one interactive local turn, which
makes the cost axes binding in practice even though no number was written down
in advance. Section~\ref{sec:rerank} is where that trade becomes decisive.

\paragraph{Ceilings.}
The current catalog limits local coverage to 450 of 600 requests. All 150
gold-delegate requests must remain delegated. Zero over-delegation therefore
gives
\[
C_{\mathrm{current}}=1-\frac{150}{600}=0.750.
\]
We label 50 delegates as external, 50 as unknown local capabilities, and 50 as
multistep. Eight multistep items contain only local steps and 42 require an
external step. Adding unknown skills and a local planner leaves $50+42=92$
blocked requests. The expanded ceiling is
\[
C_{\mathrm{expanded}}=1-\frac{92}{600}=0.847.
\]
The second ceiling depends on our delegate taxonomy. Gate and rank changes
alone cannot reach it.

\paragraph{Confirmation model.}
The policy reserves confirmation for intermediate abstention scores. It
executes below $t_{\mathrm{low}}$, delegates or
chats above $t_{\mathrm{high}}$, and asks between them. It presents the first
candidate and, after rejection, at most one alternative. A simulated user
approves the gold skill or tool and rejects every other candidate. We sweep
approval fidelity
$u\in\{1.0,0.9,0.8\}$ with fixed seed 0. The product comparison uses $u=1$.
Appendix~\ref{app:user} gives the complete rule set.

\paragraph{Repeated conditions and hardware.}
Repeated conditions separate split variation from distractor variation.
Five-fold out-of-fold gates split by skill. Candidate-set ablations use five
fixed distractor seeds and report their sample standard deviation.
Full-catalog conditions are deterministic and have no rerun variance.
Encoder scores come from multilingual-e5-base
\citep{wang2024multilinguale5} and BGE-M3 \citep{chen2024bgem3}. The lab host
has one NVIDIA L4 and four vCPUs. Appendix~\ref{app:env} fixes software
versions.

\section{Selection and Abstention}
\label{sec:different}

A 73-way nearest-item router has no representation of absence. Standalone BM25
\citep{robertson2009bm25} returns one item for every utterance, including chat
and requests that require services outside the catalog. It selected a skill
for 134 of 150 gold-delegate requests and delegated only 5 of 600. The
original single-pool ladder delegated 85 of 600 requests, below the gold rate
of 150 of 600.

Separating a catalog gate changes this failure mode. Against standalone BM25 the
gate reduces delegate-to-skill errors from 134 of 150 to 12 of 150, and mutating
false positives from 269 of 600 requests to 43 of 600. Against the single-pool
ladder it raises delegation from 85 to 141 of 600, close to the gold 150, and
end-to-end accuracy from 0.508 to 0.592. Table~\ref{tab:pool} keeps the two
baselines in separate columns. The gate does not merely adjust a confidence
threshold on the selected label. It predicts whether the selection problem is
defined.

\begin{table}[t]
\centering
\caption{The separate gate reduces unsupported local actions relative to both
single-pool baselines. The ladder applies rules, BM25, and an encoder in
sequence. All three columns score the routing decision without the confirmation
band. The same gated architecture with the band at $u=1$ is P1 in
Appendix~\ref{app:alltables}, at 0.645 accuracy.}
\label{tab:pool}
\small
\setlength{\tabcolsep}{4pt}
\begin{tabular}{@{}l S{1.3} S{1.3} S{1.3}@{}}
\toprule
Metric & \multicolumn{1}{c}{Standalone} & \multicolumn{1}{c}{Single-pool} &
\multicolumn{1}{c}{Separate} \\
 & \multicolumn{1}{c}{BM25} & \multicolumn{1}{c}{ladder} &
\multicolumn{1}{c}{gate} \\
\midrule
Accuracy & 0.367 & 0.508 & 0.592 \\
Delegate rate & 0.008 & 0.142 & 0.235 \\
Delegate to skill & \multicolumn{1}{c}{134/150} &
\multicolumn{1}{c}{54/150} & \multicolumn{1}{c}{12/150} \\
Mutating false pos. & \multicolumn{1}{c}{269/600} &
\multicolumn{1}{c}{184/600} & \multicolumn{1}{c}{43/600} \\
\bottomrule
\end{tabular}
\end{table}

All three columns score the routing decision alone, without the confirmation
band described in Section~\ref{sec:setup}. Adding the band at $u=1$ to the
gated column gives the pipeline this paper retains, P1, at 0.645 accuracy in
Appendix~\ref{app:alltables}. The 0.592 here and the 0.645 there are the same
architecture measured without and with confirmation.

This distinction resembles terminal commitment in embodied agents. Execution
and the decision to stop can diverge even when they share a trajectory
\citep{chen2026termination}. Here, retrieval and the decision to trust
retrieval diverge before execution.

\section{Selection under Matched Vocabulary}
\label{sec:selection}

\paragraph{Matched vocabulary.}
Character 3-gram BM25 over language-matched E1 documents reaches 0.988 top-1
on match utterances. Its serialized index is 17 KB and its CPU median latency
is 0.140 ms. At the study's 50:50 lexical mixture, standalone BM25 reaches
0.750, e5 reaches 0.780, and BGE reaches 0.843
(Table~\ref{tab:selection}). The aggregate ordering hides where BM25 succeeds.

\begin{table}[t]
\centering
\caption{Matched vocabulary nearly removes BM25 selection error, while the
50:50 mixture favors the encoders.}
\label{tab:selection}
\begin{tabular}{l S{1.3} S{1.3} S{1.3}}
\toprule
Ranker & \multicolumn{1}{c}{Match} & \multicolumn{1}{c}{Mismatch} &
\multicolumn{1}{c}{50:50 mix} \\
\midrule
BM25+E1 & 0.988 & 0.512 & 0.750 \\
e5+E0 & 0.951 & 0.608 & 0.780 \\
BGE+E1 & 0.957 & 0.729 & 0.843 \\
\bottomrule
\end{tabular}
\end{table}

\paragraph{Alias controls.}
Alias content improves lexical retrieval rather than merely lengthening the
catalog. The base index, E0, uses names, descriptions, and slot names. The
alias-expanded index, E1, adds aliases authored from those fields without
opening utterance text. E1 raises mismatch accuracy from 0.380 to 0.512.
Two length controls bracket that gain, and they pad to different targets.
Matching E1 exactly, by padding each E0 document with function words to that
document's E1 character 3-gram count, gives 0.349 at a mean of 102.5 3-grams.
Padding E0 by a fraction of its own token count instead gives 0.361, 0.380,
0.373, and 0.367 at 0.25, 0.5, 1, and 2 times, where the last reaches a mean of
165.3 3-grams and so overshoots E1. Both controls stay near E0 on either side of
E1's length. Shuffling alias lists among catalog items gives 0.127. With BM25
length normalization disabled, $b=0$, E1 retains a 0.151 gain. It reaches 0.512
versus 0.361, and its top-1 accuracy is unchanged at $b=0$, $0.75$, and $1$,
so the gain is not an artifact of length normalization. These are confirmatory
controls for lexical content, not additional catalog searches.

Examples do not improve the alias-expanded index. The example-expanded index,
E2, reaches 0.512 mismatch accuracy. Its length-matched filler control reaches
0.518 in the final recheck. We therefore reject E2 and keep E1. The catalog
edit repairs lexical retrieval rather than the encoder.

The two encoders respond differently to aliases, and both differences are a few
items. Aliases move BGE mismatch by three items, from 0.711 to 0.729, and its
match by one item, from 0.963 to 0.957. On e5 the mismatch count is unchanged at
101 of 166 while match falls by two items, from 0.951 to 0.939. None of these
differences is separated from zero at this sample size. We nonetheless report
each ranker on the index that scores higher for it, e5 on E0 and BGE on E1,
which is worth 0.006 at the study's mixture in both cases, from 0.774 to 0.780
and from 0.837 to 0.843. That choice is mildly optimistic for the encoders, and
the crossover we compute for BGE in Section~\ref{sec:choice} inherits it.

\begin{figure}[!htbp]
\centering
\includegraphics[width=\columnwidth]{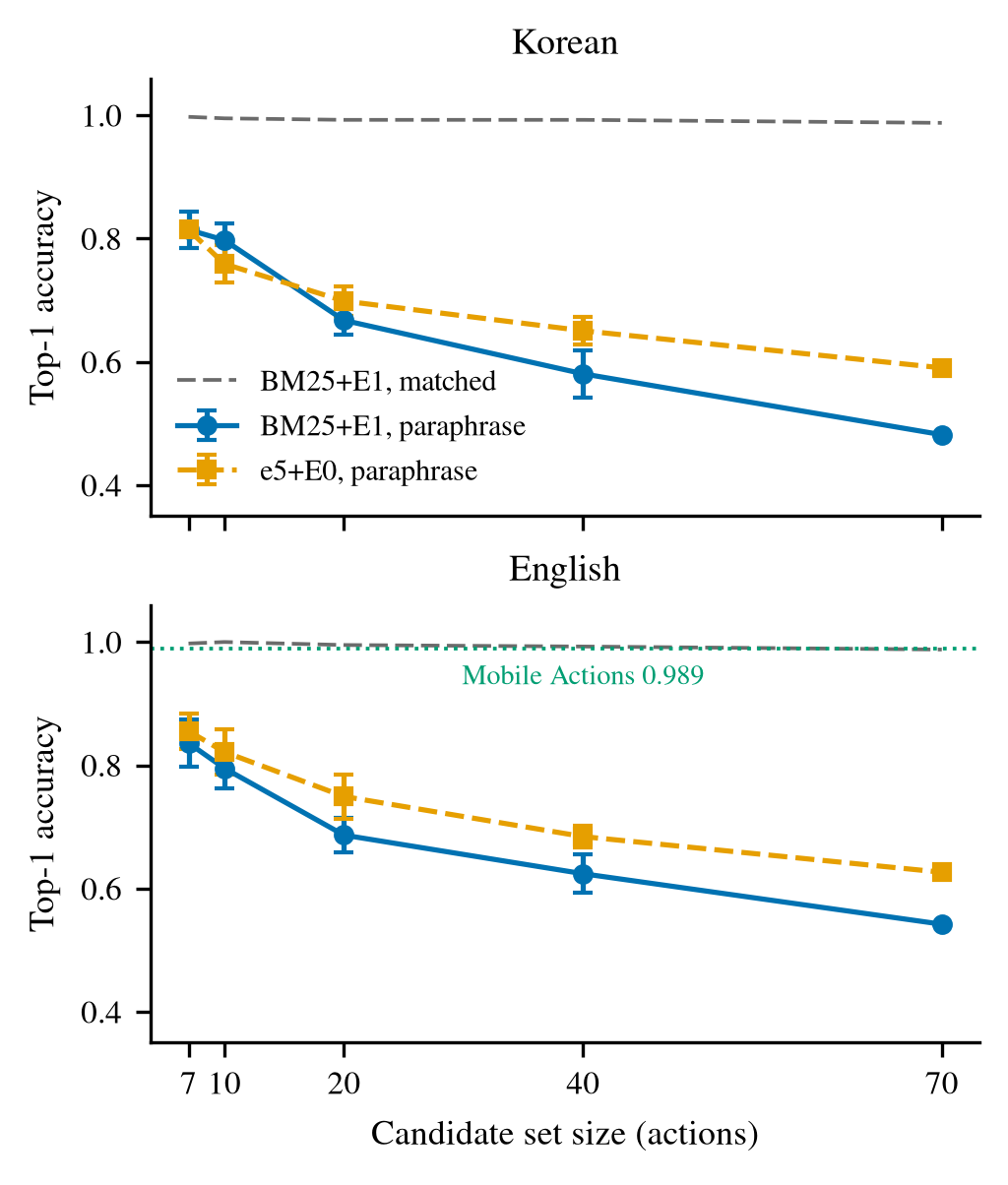}
\caption{Candidate restriction helps both rankers on paraphrases but never
reaches the Mobile Actions line, which instead sits inside the range of our own
matched values, 0.988 to 1.000 in English.
Section~\ref{sec:choice} reads the external number against the matched side of
the split for that reason. The matched curve is drawn for BM25 only, since the
external comparison concerns the lexical ranker. Error bars show one sample
standard deviation over five distractor seeds.}
\label{fig:candidates}
\end{figure}

\paragraph{Candidate count.}
The advantage also depends on the active set. We construct candidate sets that
always contain gold and sample the remaining tools uniformly. Over five
distractor seeds, BM25 mismatch rises from 0.512 at 70 candidates to
$0.825\pm0.023$ at seven. e5 rises from 0.608 to $0.835\pm0.016$. At seven and
ten candidates the seed spread exceeds the gap between the two rankers, and the
per-seed winner changes: BM25 leads on two of five seeds at seven candidates and
on three of five at ten. From 20 candidates upward e5 leads on all five seeds.
At ten candidates BM25 is not separated from e5 on either lexical slice. The encoder
advantage appears clearly at 20, 40, and 70 candidates, not at ten or fewer.
Table~\ref{tab:candidates} and Figure~\ref{fig:candidates} provide the full
curve.

\begin{table*}[t]
\centering
\caption{Candidate restriction narrows the mismatch-selection difference.
Values are mean top-1 accuracy and sample standard deviation over five
distractor seeds. The full 70-document pool is deterministic.}
\label{tab:candidates}
\small
\begin{tabular}{S{2.0} *{8}{S{1.3}}}
\toprule
\multicolumn{1}{c}{$k$} &
\multicolumn{2}{c}{BM25 match} & \multicolumn{2}{c}{BM25 mismatch} &
\multicolumn{2}{c}{e5 match} & \multicolumn{2}{c}{e5 mismatch} \\
\cmidrule(lr){2-3}\cmidrule(lr){4-5}\cmidrule(lr){6-7}\cmidrule(lr){8-9}
& \multicolumn{1}{c}{Mean} & \multicolumn{1}{c}{SD} &
\multicolumn{1}{c}{Mean} & \multicolumn{1}{c}{SD} &
\multicolumn{1}{c}{Mean} & \multicolumn{1}{c}{SD} &
\multicolumn{1}{c}{Mean} & \multicolumn{1}{c}{SD} \\
\midrule
7 & 0.998 & 0.003 & 0.825 & 0.023 & 0.995 & 0.005 & 0.835 & 0.016 \\
10 & 0.998 & 0.003 & 0.796 & 0.029 & 0.989 & 0.007 & 0.790 & 0.026 \\
20 & 0.994 & 0.000 & 0.677 & 0.017 & 0.979 & 0.005 & 0.724 & 0.016 \\
40 & 0.993 & 0.005 & 0.602 & 0.029 & 0.960 & 0.005 & 0.667 & 0.020 \\
70 & 0.988 & 0.000 & 0.512 & 0.000 & 0.951 & 0.000 & 0.608 & 0.000 \\
\bottomrule
\end{tabular}
\end{table*}

\section{Abstention from BM25 Scores}
\label{sec:abstention}

Strong matched-vocabulary selection does not produce a reliable out-of-catalog
signal. On mismatch-in versus out, the six BM25 score features range from
0.480 to 0.610 AUC individually. Score per term reaches 0.480, normalized margin
0.542, and score over mean 0.550. Raw margin reaches 0.569, top-k entropy 0.587,
and top score 0.610.

Combining BM25 score features still leaves the difficult subset unresolved.
The combined score gate, G3, is a balanced logistic regression over seven
features plus an intercept. Its eight-parameter coefficient payload occupies
454 bytes. Five-fold out-of-fold AUC is 0.831 overall and 0.959 on match-in
versus out. It reaches only 0.697 on mismatch-in versus out.
Table~\ref{tab:gates} shows the comparison.

The seventh feature deserves its own account, because alone it beats the
combination. Query character 3-gram count fitted out of fold on its own reaches
0.741, above G3's 0.697, and as an untransformed feature it reaches 0.743. G3
with the feature removed falls to 0.587. Query length therefore carries most of
what G3 has, and adding the six score features to it costs out-of-fold AUC
rather than adding to it. We do not read 0.741 as a routing signal. Length
separates our authored sets partly by construction, since mismatch items were
written as paraphrases and out-of-catalog items as chat or multistep requests.
The conclusion survives either reading: taken as an optimistic bound it still
leaves every signal on this side of the comparison below the 0.806 encoder
gate.

\begin{table}[t]
\centering
\caption{Encoder similarities separate paraphrases from out-of-catalog requests
better than any lexical or query-statistic signal we measured. Every trained
row is out of fold. The encoder rows are frozen similarities with no training
on this task. The comparison uses $n_{\mathrm{in}}=166$ and
$n_{\mathrm{out}}=210$.}
\label{tab:gates}
\small
\begin{tabular}{@{}l S{1.3}@{}}
\toprule
Gate signal & \multicolumn{1}{c}{AUC} \\
\midrule
Six score features, individually & \multicolumn{1}{c}{0.480--0.610} \\
Query length alone & 0.741 \\
G3 without query length & 0.587 \\
G3, all seven features & 0.697 \\
BGE-M3 gate & 0.803 \\
e5-base gate & 0.806 \\
\bottomrule
\end{tabular}
\end{table}

Frozen encoder similarities separate mismatch-in from out more effectively.
e5 reaches 0.806 mismatch AUC and BGE reaches 0.803. The e5 ONNX int8 artifact
uses 278 MB and takes 13.7 ms CPU median latency. The selected split pipeline,
P1, therefore spends the neural model on abstention. It leaves selection to
the 17 KB lexical index. This result does not show that every gate requires an
encoder. It shows that the measured BM25 scores do not resolve this gate.

The gap that decides the architecture survives resampling, and the gap between
the two encoders does not. On a cluster bootstrap over utterance identifiers
with 2000 resamples, e5 exceeds G3 by 0.109 AUC with a 95\% interval of
$[0.059,0.159]$, and exceeds query length alone by 0.064 with interval
$[0.014,0.117]$; the displayed three-decimal values give 0.065 for that second
difference. BGE exceeds G3 by 0.106 with interval $[0.055,0.153]$. e5 and
BGE differ by 0.003 with interval $[-0.027,0.032]$, so the two encoders are not
separated on this slice. Adding query length to the six score features moves G3
by $+0.110$ with interval $[0.073,0.147]$, which confirms how much of G3 that
one feature carries. The reverse direction, adding the six features to query
length alone, costs 0.044. Resampling paired Korean and
English items together instead of single utterances moves every bound by at most
0.007 and changes no conclusion about whether an interval spans zero. The
reported architecture claim is therefore that
both frozen encoders beat every BM25-side signal we measured, not that one
encoder beats the other.

\begin{figure}[!htbp]
\centering
\includegraphics[width=\columnwidth]{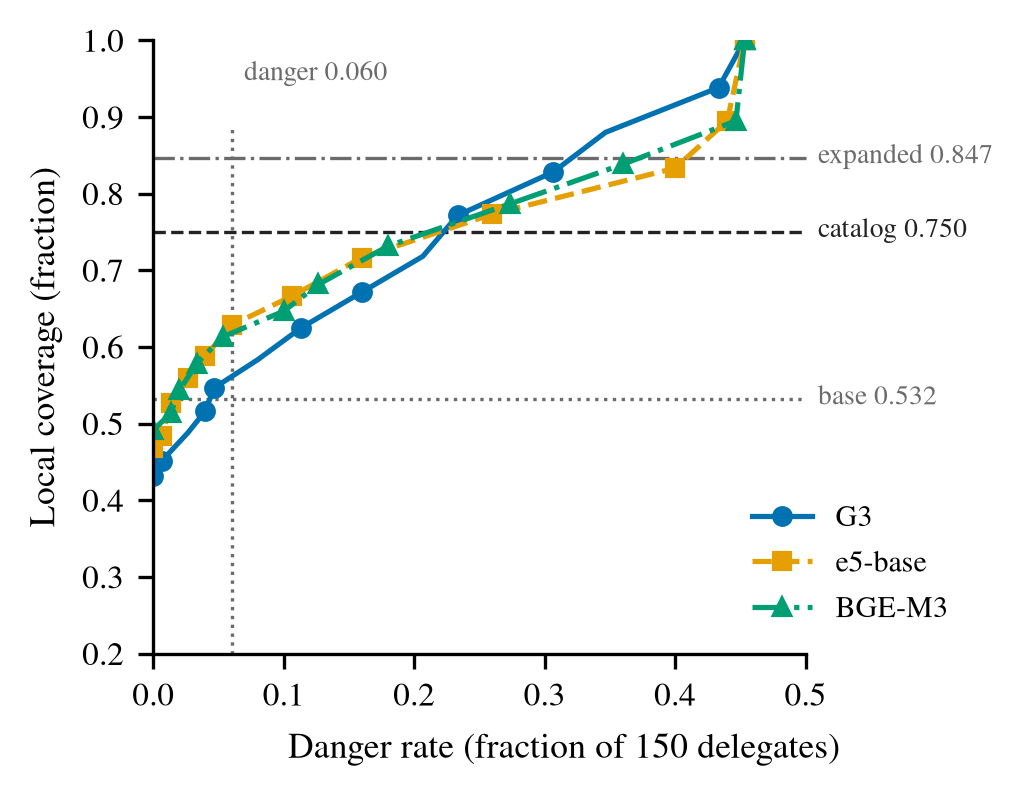}
\caption{Both encoder gates hold more local coverage than the BM25-score gate
throughout the danger budget. Reference lines mark the base rate and the two
catalog ceilings.}
\label{fig:pareto}
\end{figure}

That AUC ordering carries over to the operating points a product must choose
from. Figure~\ref{fig:pareto} shows that both encoder gates deliver more local
coverage than the BM25-score gate at every danger rate within the 0.060 budget.
The BM25-score gate passes both encoders only above a danger rate of 0.233,
which is roughly four times the budget.
The same axes carry both catalog ceilings, so the figure also shows that the
best gate reaches 0.628 coverage inside the budget, well below the 0.750
current-catalog ceiling. Approaching either ceiling requires danger rates that
the product rule excludes.

\begin{figure}[!htbp]
\centering
\includegraphics[width=\columnwidth]{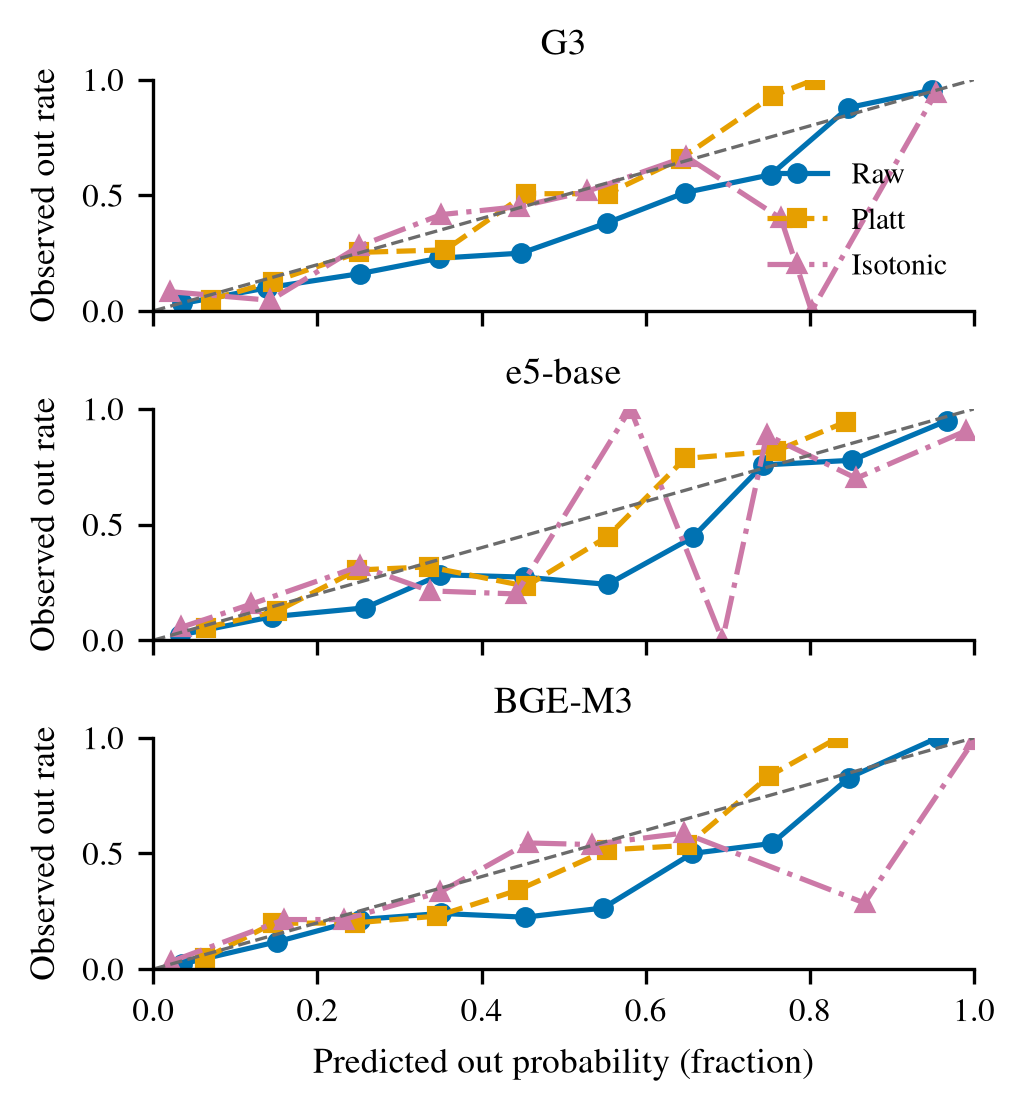}
\caption{Calibration moves predicted probabilities toward the diagonal but
does not change ranking. The staircase in the isotonic curves is an
overfitting artifact of the finite calibration folds.}
\label{fig:reliability}
\end{figure}

Calibration does not rescue the BM25-score gate, because it applies a monotone
score transformation. Platt scaling lowers all-set expected calibration error
from 0.101 to 0.041, but G3 mismatch AUC changes only from 0.697 to 0.696 at
the product operating point. Isotonic calibration yields 0.685 there. More
importantly, raw mismatch maximum calibration error is 0.922. A confirmation
band requires probabilities whose intervals have stable empirical meaning.
This error prevents a narrow BM25-score band from serving that role.
Figure~\ref{fig:reliability} shows the reliability curves.

\section{Reranking and Neural Ranking}
\label{sec:rerank}

\paragraph{Reranking the shortlist.}
BM25 mismatch recall rises from 0.512 at rank 1 to 0.741 at rank 3, 0.819 at
rank 5, and 0.922 at rank 20. A large fraction of gold items is present but
misordered. This motivates a smaller second-stage problem. The first stage
retrieves five with BM25 E1. The second stage scores only those five.

Table~\ref{tab:rerank} rejects a simple version of that proposal. e5 top-5 reranking
reaches 0.602 mismatch, slightly below its 0.608 full 70-document selection.
BGE reaches 0.699. Relative to the interval from BM25 top-1 0.512 to the
top-5 ceiling 0.819, e5 recovers 29.4\% and BGE recovers 60.8\%. The existing
Qwen3-4B top-5 run reaches 0.777 and recovers 86.3\%.

\begin{table}[t]
\centering
\caption{Native tool calling recovers more of the BM25 shortlist advantage
than either encoder reranker on mismatch requests.}
\label{tab:rerank}
\small
\begin{tabular}{S{1.0} l S{1.3} S{1.3} S{1.3}}
\toprule
\multicolumn{1}{c}{$k$} & Reranker & \multicolumn{1}{c}{Top-1} &
\multicolumn{1}{c}{Recall@$k$} & \multicolumn{1}{c}{Recovery} \\
\midrule
5 & none & 0.512 & 0.819 & 0.000 \\
5 & e5 & 0.602 & 0.819 & 0.294 \\
5 & BGE & 0.699 & 0.819 & 0.608 \\
5 & Qwen3-4B & 0.777 & 0.819 & 0.863 \\
3 & e5 & 0.608 & 0.741 & 0.421 \\
3 & BGE & 0.669 & 0.741 & 0.684 \\
\bottomrule
\end{tabular}

\smallskip
{\footnotesize The two encoders rerank on CPU. Qwen3-4B is a native tool caller
on GPU, not a reranker, and is listed here because it consumes the same
five-candidate shortlist.}
\end{table}

The encoder's errors are not limited to gold being displaced from first to
positions two through five. When the gold item receives low semantic score,
removing 65 other candidates does not increase that score relative to the
remaining confounders. The e5 top-3 result equals full-pool e5 at 0.608. This
supports the narrower interpretation. It does not prove that every encoder
reranker has the same error.

The reranked system does not replace the selected split architecture. P5
combines BM25 top-5, e5 reranking, and the same e5 E0 gate. It reaches 0.642
accuracy, 0.763 completion, 0.615 coverage, and 0.060 danger. P1 reaches 0.645,
0.755, 0.627, and 0.060. A candidate displaces the retained pipeline only if it
is at least as good on accuracy, coverage, and danger and strictly better on at
least one, a rule recorded with the round-4 results rather than in the
preregistration. We call it the adoption rule below. Completion is not one of
its three metrics because it credits a safe delegation as well as a local
completion, so it rises when a pipeline delegates more. P5 is the case in point.
It completes five more requests than P1 while resolving two fewer exactly, 385
against 387, and it delegates seven more. It is lower on accuracy and on
coverage, so it does not qualify and P1 stays selected.

\paragraph{Replacing the ranker.}
Neural ranking does improve routing quality, and the reason we do not adopt it
is cost. This condition does not use a shortlist at all: P3 replaces the BM25
ranker with BGE over the full pool and keeps the e5 gate. It reaches
0.660 accuracy, 0.768 completion, and 0.628 coverage against P1's 0.645, 0.755,
and 0.627, at the same 0.060 danger and the same 0.198 ask rate. P3 therefore
satisfies the adoption rule, and what follows rejects it on cost rather than on a
metric. Counted in
requests, P3 resolves nine more than P1 on accuracy and eight more on
completion, and keeps one more request local, 377 of 600 against 376. It also
delegates one request fewer, so neither margin comes from the delegation credit.
The two
larger margins are what the comparison rests on. Its CPU median is 197.3 ms
against 13.9 ms, a factor of 14, and it needs 846.1 MB across two models rather
than 278.3 MB across one. Adopting P3 would pay 14 times the latency and three
times the memory for nine requests of accuracy, eight of completion, and one of
coverage. We decline that trade and retain P1.
Appendix~\ref{app:alltables} lists both pipelines with their cost columns.

\begin{table}[t]
\centering
\caption{No alternative replaces P1. P5 is rejected on quality at equal cost;
the BGE ranker and the native caller exceed P1 on quality and are rejected on
measured cost. All four rows hold danger at 0.060.}
\label{tab:caller}
\small
\begin{tabular}{@{}l S{1.3} S{1.3} r@{}}
\toprule
System & \multicolumn{1}{c}{Mism.} & \multicolumn{1}{c}{Cov.} &
\multicolumn{1}{c}{Latency} \\
\midrule
P1, BM25 rank & 0.512 & 0.627 & 13.9 CPU \\
P5, e5 rerank & 0.602 & 0.615 & 13.9 CPU \\
P3, BGE rank & 0.729 & 0.628 & 197.3 CPU \\
Qwen3-4B top-5 & 0.777 & 0.645 & 1,111 GPU \\
\bottomrule
\end{tabular}

\smallskip
{\footnotesize Mism.\ is mismatch top-1 accuracy and Cov.\ is local coverage.
Latency is the median in milliseconds on the stated processor. All three CPU
systems use the same e5 gate. They differ only in the ranker.}
\end{table}

The small language model (SLM) adds eleven requests of coverage over P1 and
eighteen over P5, at about 80 times the P1 CPU median.
Table~\ref{tab:caller} keeps hardware labels because GPU and CPU latency are
not interchangeable.

Prefix caching does not reverse this comparison. In the full-tool condition,
599 of 600 prompts reuse the exact tool prefix and \code{vLLM} reports 99.4\% prefix
cache hit rate. Under top-5, exact prefix reuse falls to 5.8\% because candidate
sets vary. Mean prompt tokens fall from 7,987 to 916, yet Qwen median latency
improves only 41 ms, from 1,152 to 1,111 ms. On this GPU and runtime, the loss
of prefix reuse offsets little of the token reduction.

\section{Checkpoint Deployment}
\label{sec:deploy}

The native tool-call roster contains four checkpoints. Two produce 600 rows in
both full and top-5 conditions. Two produce no routing rows under the fixed
\code{vLLM} 0.27.1 protocol. Table~\ref{tab:deployment} reports exclusions as deployment
outcomes, not accuracy results.

\begin{table*}[t]
\centering
\caption{Only two of four fixed checkpoints produce routing rows under the
specified runtime and native tool-call protocol.}
\label{tab:deployment}
\small
\begin{tabularx}{\textwidth}{l l Y}
\toprule
Checkpoint & Status & Measured condition or blocker \\
\midrule
LFM2.5-2.6B & measured & full and top-5, 600 each \\
Qwen3-4B & measured & full and top-5, 600 each \\
Gemma 4 E4B-it & load failed & heterogeneous \code{head\_dim} conversion \\
Granite 4.0 H-Small & excluded & required Pythonic parser absent \\
\bottomrule
\end{tabularx}
\end{table*}

Gemma 4 E4B-it fails before weight admission. \code{vLLM} reads a global
\code{head\_dim} from a heterogeneous per-layer configuration and raises
\code{AmbiguousGlobal}\allowbreak\code{PerLayer}\allowbreak\code{AttributeError}.
Granite 4.0 H-Small requires a
Python-style tool parser. \code{vLLM} 0.27.1 provides \code{granite} and
\code{granite4}, but not \code{granite\_pythonic}. Its BF16 checkpoint also
exceeds the L4's 23,034 MiB, and we do not substitute a quantized checkpoint.
Appendix~\ref{app:errors} includes the captured errors and environment.

LFM2.5-2.6B completes both conditions. With
\code{tool\_choice="required"}, its full-catalog no-tool-call rate is 14.3\%,
86 of 600. Its top-5 rate is 9.8\%, 59 of 600. This is a protocol outcome, not
a judgment about other runtimes.
Needle 2 separates the same two decisions in its architecture
(\cref{sec:intro}). An independent C engine for it reports 90.8\%
tool-name accuracy on Mobile Actions against 98.1\% for the official engine, and
lists as a limitation that its own build ``will not decline'' because it skips
the probe heads carrying those weights \citep{mimimodel2026}. What the
reimplementation drops is the gate and not the ranker, which is the same
asymmetry we measure: selection transfers to a compact substitute, abstention
does not. Both are vendor and repository reports, not reruns on our catalog.

\section{Ranker Choice}
\label{sec:choice}

Vocabulary match and candidate count jointly determine ranker choice. Let $p$
be the match share. At 70 candidates, weighted selection is
$p a_{\mathrm{match}}+(1-p)a_{\mathrm{mismatch}}$. BM25 exceeds e5 above
$p=0.725$ and BGE above $p=0.877$. These are not catalog-independent
constants. Against e5, the measured crossover is 0.798 at seven candidates.
At ten candidates the five-seed means do not cross inside $[0,1]$, so the two
rankers are not separated at that size. The crossover is 0.763 at 20 candidates,
0.664 at 40, and 0.725 at 70 (Table~\ref{tab:crossover}).

Two properties of that column limit how far it can be read. The denominator is
the mismatch gap itself, so where the gap is smaller than the seed standard
deviation the value is numerically unstable. That holds at seven and ten
candidates, where the gaps are 0.010 and 0.006 against standard deviations of
0.016 to 0.029. At seven candidates the denominator is 0.013, so recomputing
that row from the three-decimal values printed in the table shifts it by about
0.03. The column is also not monotone in $k$: it falls from 0.763 at
20 candidates to 0.664 at 40 and then rises to 0.725 at 70. We therefore read
the column as evidence that the crossover moves with candidate count, not as a
calibrated function of it.

\begin{table}[t]
\centering
\caption{The match share needed for BM25 to exceed e5 changes with candidate
count. BM25 and e5 columns report mismatch accuracy. Crossovers use unrounded
five-seed means, not the displayed values, and the five-seed means do not cross
at ten candidates.}
\label{tab:crossover}
\small
\begin{tabular}{S{2.0} S{1.3} S{1.3} S{2.3}}
\toprule
\multicolumn{1}{c}{$k$} & \multicolumn{1}{c}{BM25} &
\multicolumn{1}{c}{e5} & \multicolumn{1}{c}{Crossover $p$} \\
\midrule
7 & 0.825 & 0.835 & 0.798 \\
10 & 0.796 & 0.790 & \multicolumn{1}{c}{none (means)} \\
20 & 0.677 & 0.724 & 0.763 \\
40 & 0.602 & 0.667 & 0.664 \\
70 & 0.512 & 0.608 & 0.725 \\
\bottomrule
\end{tabular}
\end{table}

\begin{figure*}[!t]
\centering
\includegraphics[width=\textwidth]{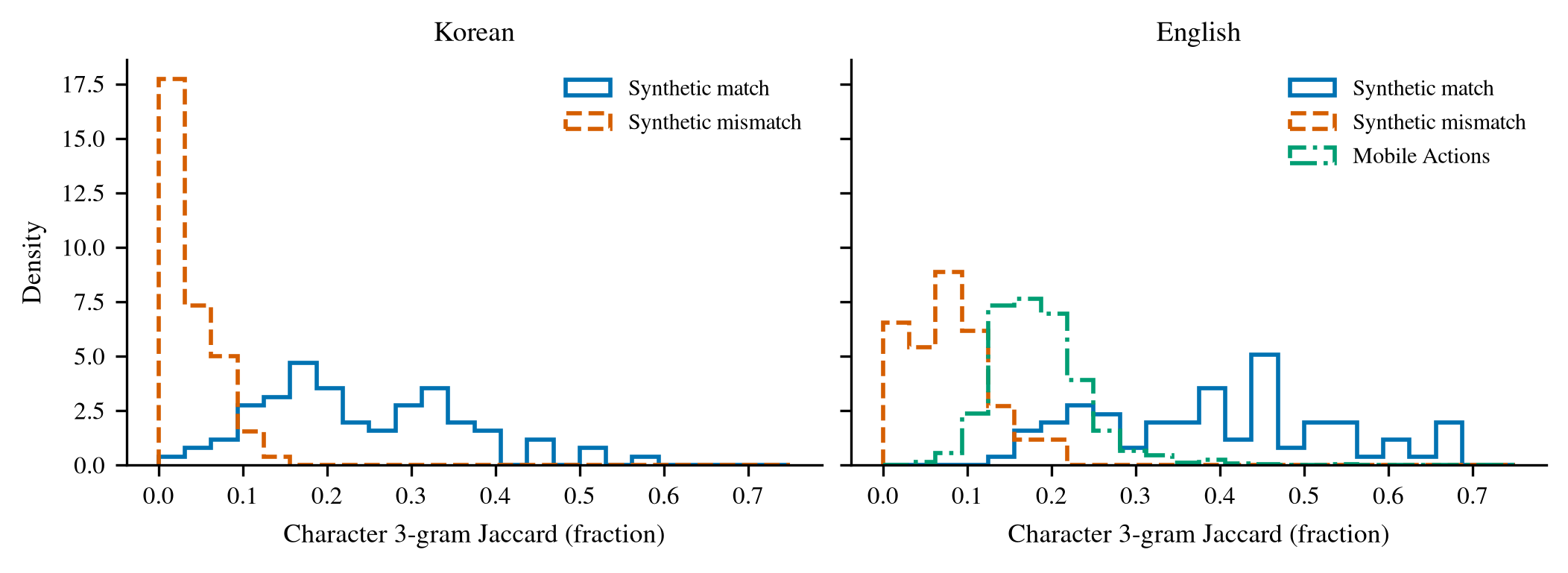}
\caption{Mobile Actions requests share more character 3-grams with their
catalog than synthetic mismatch requests do with theirs, but fewer than our
matched English slice, even though their accuracy sits at the matched level.
Jaccard is a vocabulary proxy, not a causal decomposition, and it underpredicts
the Mobile Actions score.}
\label{fig:jaccard}
\end{figure*}

The external cross-check sits above our paraphrase slice, and comparing it only
against that slice overstates the gap. Mobile
Actions \citep{mobileactions2025} supplies English-only records with seven
candidates each. BM25 tool-name top-1 is 0.989. That number cannot be compared
fairly with a mixed Korean-English ablation.
The language split leaves most of that difference unexplained. Synthetic
English mismatch reaches only 0.836 at seven candidates. Korean reaches 0.814.
English therefore does not explain the difference.

Mobile Actions describes its actions in words close to the commands that invoke
them, which places it on the matched-vocabulary side of our split. Holding language
fixed at seven candidates, our English matched slice reaches 0.998 and our
English paraphrase slice 0.836. For BM25 at this candidate count the Korean
matched value is identical to the English one, so the pooled matched figure in
Table~\ref{tab:candidates} introduces no language mixing at this comparison
point. Mobile Actions sits inside that interval and near its matched end. The external set is consistent with our matched-vocabulary result
rather than above it.

Character 3-gram Jaccard also changes by language. Synthetic English match has
median 0.401 and interquartile range 0.261 to 0.491. English mismatch has
median 0.079 and range 0.035 to 0.101. Mobile Actions has median 0.180 and
range 0.148 to 0.212. Korean match and mismatch medians are 0.217 and 0.029.
Figure~\ref{fig:jaccard} compares Mobile Actions only with the English panel.

The proxy and the accuracy disagree, which is itself the argument against a
single-variable account. Mobile Actions has a lower median Jaccard than our
matched English slice, 0.180 against 0.401, yet its accuracy sits at the matched
level rather than the paraphrase level. Vocabulary overlap measured this way
therefore underpredicts its score, so we cannot attribute 0.989 to overlap
alone.

Neither candidate count nor language alone explains 0.989. The external set
combines seven candidates, English commands, and descriptions close to those
commands. These conditions overlap. We therefore do not attribute its score
to one variable. The measured outcome depends on match share and active
candidate count, with language-specific text statistics.
Neither production match share nor production active-set size is observed in
this study.

\section{Conclusion}

Only one of the router's two decisions is a retrieval problem. A character
3-gram index selects reliably when the request and the catalog share
vocabulary, and restricting the active candidate set recovers most of what
paraphrasing costs it. No feature derived from that index's own scores yields
a reliable in-catalog gate, and improving selection does not produce one.
Closing the gate requires a frozen encoder, and abstention rather than ranking
is therefore the measured justification for a neural component on the device.

The choice of ranker is a separate decision, and it resolved in the opposite
direction. The encoder reranker is rejected on quality at equal cost, running
at the same 13.9 ms median in the same 278.3 MB and one model. The BGE ranker and the
native caller are rejected on cost while beating the retained pipeline on
quality, at 14 and about 80 times that median. All three do recover part of the
difference between top-1 accuracy and top-5 recall, 29.4\% with e5 reranking,
60.8\% with BGE reranking and 86.3\% with the 4-billion-parameter native caller,
and none of them enters the retained pipeline. Half the fixed checkpoint roster
produces no routing rows at all.

The architecture is therefore determined by abstention and the ranker by a cost
budget. A deployment with a larger latency and memory budget should prefer the
neural ranker, a choice this study can inform but cannot make, since production
lexical match share and active candidate count are unobserved here.

The next measurable questions are distributional. A production trace can
estimate active-set size, lexical match share, and confirmation acceptance
without changing catalog contracts. A hardware replication can test whether
the latency ordering survives an 8 GB GPU. A multi-step extension can measure
whether independent gate and execution errors compose geometrically or show
the correlated decay reported in long-horizon agents
\citep{khanal2026reliability}.

\section*{Limitations}
\label{sec:limits}
\addcontentsline{toc}{section}{Limitations}

The 600 utterances are synthetic. We set the 50:50 match/mismatch mixture and
25\% delegate rate. The 70-document catalog is also our choice. Candidate
ablation measures sensitivity, but it samples distractors uniformly rather
than recovering a production active set.

None of the three thresholds fixed in advance was a cost limit, so rejecting P3
at 197.3 ms rests on a
budget we state after seeing the measurement rather than before. A reader with a
larger latency and memory budget should read P3, not P1, as the better pipeline
in Table~\ref{tab:pipelines}. Its coverage advantage is one request out of 600, so the
comparison rests on the nine-request accuracy margin instead. The preregistered coverage target has a related looseness. It
was set at 0.628, the coverage of an e5 reference cell whose ask rate is 0.415
and which therefore violates the 0.200 ask cap the same preregistration
imposes. Under that cap the comparable e5 baseline is 364 of 600 requests, and
P1 clears it by twelve requests at 376. The preregistered target of 377 is one
request above P1 and exactly P3's count. Three
different quantities in this paper round to 0.628: that preregistered target,
P3's pipeline coverage, and the best gate-sweep coverage inside the danger
budget in Figure~\ref{fig:pareto}. The coincidence is numerical only.

Mobile Actions combines seven candidates, English, and descriptions close to
commands. The cross-check cannot attribute its score to one factor.
Language-specific Jaccard plots reduce one comparability error but do not make
the catalogs or task semantics identical.

Each ranker is reported on the catalog index that suits it, BM25 and BGE on E1
and e5 on E0. That is mildly optimistic for the encoders, by 0.006 at the
study's mixture for each of them, and the BGE crossover carries the same choice.

The user-response model is simulated. Approval fidelity 1.0, 0.9, and 0.8 is a
sweep, not observed behavior. The 20\% ask cap is a policy value. We do not
measure patience. The 0.750 and 0.847 ceilings depend on our delegate taxonomy.

The preregistered signal-diversity hypothesis is rejected. Mean
shared-minus-split MCC is $+0.062$ with bootstrap 95\% interval
$[-0.013,+0.140]$.
P2 uses e5 for both ranking and gating. P3 uses BGE for ranking and e5 for
gating. Ask-band top-1 follows the same order as coverage across P3, P1, and
P2. The three ask-band accuracies are 0.421, 0.368, and 0.342. Their coverage
values are 0.628, 0.627, and 0.607. We report association, not causation. The
e5-only system also uses a narrower $t_{\mathrm{high}}$. Its confirmation band
contains 38 of 600 requests, compared with 57 of 600 for P1.

The fixed protocol does not measure two of four SLM checkpoints. The roster
comparison is therefore incomplete. All latency measurements use one NVIDIA
L4 and four vCPUs. We do not reproduce them on an 8 GB consumer GPU. Longer
tool trajectories can be less reliable than independent per-step estimates
predict \citep{khanal2026reliability}. Our single-decision experiment cannot
estimate that decay.

The evaluation excludes argument extraction. We score tool-name routing and
deterministic post-state completion. We cannot report strict exact tool-call accuracy
against benchmarks that include argument values. FunctionChat-Bench's dialog
dimensions therefore remain related context, not a merged score
\citep{lee2024functionchat}.

\section*{Ethics Statement}
\addcontentsline{toc}{section}{Ethics Statement}

The 600 utterances were written by the author for this study. They contain no
personal data, no real user traffic, and no content collected from a person.
The catalog is a specification of local actions on a device and carries no user
state. Neither the utterances nor the catalog is public at the time of writing. No human subjects took part. The confirmation experiment replaces a
user with a rule that approves the gold action and rejects every other
candidate, and the paper reports it as a simulation rather than as observed
behavior.

The routing decision this paper studies is a safety decision as much as a
quality one. A router that selects a local action when none applies executes
something the user did not ask for, and 51 of the 70 searchable actions in
this catalog change state on the device rather than only reading it. That is
why the danger rate counts gold-delegate requests sent to any skill or file
tool, why it is capped in advance rather than traded against coverage, and why
every pipeline in this paper is reported at the same danger budget. A reader who adopts the retained pipeline inherits
that cap rather than the accuracy number alone.

The two encoder checkpoints are used as released and neither is fine-tuned
here, so nothing in this work redistributes a modified model.
Appendix~\ref{app:env} records what each snapshot states about its own licence,
including that the multilingual-e5-base snapshot we pulled carries neither a
licence file nor a licence line in its model card.

\bibliography{refs}

\clearpage
\appendix
\setcounter{table}{0}
\renewcommand{\thetable}{A\arabic{table}}
\setcounter{figure}{0}
\renewcommand{\thefigure}{A\arabic{figure}}

\section{Environment}
\label{app:env}

Table~\ref{tab:environment} lists the host, accelerator, and library versions
behind every latency and deployment result in the paper. One host served all
conditions, so no result mixes hardware. Its last two rows record what each
encoder snapshot states about its own licence, read from the snapshot rather
than from the model page, which is why one of them is blank rather than MIT.

\begin{table}[!ht]
\centering
\caption{All latency and deployment results use the same fixed host.}
\label{tab:environment}
\small
\begin{tabularx}{\columnwidth}{@{}l Y@{}}
\toprule
Item & Value \\
\midrule
Host & Linux, Python 3.12.3, 4 vCPU \\
GPU & NVIDIA L4, 23,034 MiB, driver 595.91.07 \\
\code{vLLM} & \code{0.27.1} \\
\code{torch} & \code{2.13.0+cu130} \\
CUDA from \code{torch} & \code{13.0} \\
\code{transformers} & \code{5.15.1} \\
Encoder runtime & ONNX Runtime CPU, int8 for e5 \\
BGE-M3 licence & MIT, from the model card in the snapshot \\
e5 licence & no licence file and no card line in the snapshot \\
\bottomrule
\end{tabularx}
\end{table}

\section{Nine Non-Skill Action Specifications}
\label{app:tools}

Table~\ref{tab:actions} gives the identifier, name, retrieval description, and
required slots for the six file tools and the three meta actions. The 64 skills
follow the same document and contract format and are not listed individually.

\begin{table*}[!t]
\centering
\caption{The six file tools and three meta actions complete the action space.
The last column names each required slot and its type.}
\label{tab:actions}
\small
\begin{tabularx}{\textwidth}{l l Y l}
\toprule
ID & Name & Description & Required slots \\
\midrule
\code{fs.list} & list & List file and subdirectory names & path:dir \\
\code{fs.read} & read & Return file contents unchanged & path:file \\
\code{fs.write} & write & Create or overwrite a file & path:file, content:string \\
\code{fs.edit} & edit & Replace a specified text span & path:file, old:string, new:string \\
\code{grep} & grep & Find regex-matching lines across files & path:dir, pattern:string \\
\code{table.query} & table query & Run a SQL-like filter over a table & path:file, query:string \\
\code{meta\_chat} & chat & Greeting, small talk, or opinion & none \\
\code{meta\_ask} & ask & Clear task with a missing required value & missing:string \\
\code{meta\_delegate} & delegate & External service or unsupported multistep task & none \\
\bottomrule
\end{tabularx}
\end{table*}

\section{Catalog Statistics}
\label{app:catalog}

The retained catalog index expands each document with authored aliases. E1
indexes 70 non-meta documents separately by language. Index text joins
name, description, aliases, and slot names. Korean character length has mean
73.0, median 73, range 46 to 89, and interquartile range 67 to 80. English has
mean 136.0, median 136, range 101 to 171, and interquartile range 123.8 to
148.8. E0 mean character 3-gram count is 55.1 and E1 is 102.5 over
70 documents times two languages. The E1-filler control matches 102.5 exactly.

\section{Gate Features}
\label{app:features}

The combined score gate uses seven signals from each ranked score list. For
ranked scores $s_1\geq s_2\geq\cdots$, G3 uses
\begin{enumerate}
  \item top score $s_1$,
  \item margin $s_1-s_2$,
  \item normalized margin $(s_1-s_2)/(|s_1|+\epsilon)$,
  \item entropy of the normalized top-k score distribution,
  \item $s_1$ divided by mean top-k score,
  \item query character 3-gram count, and
  \item $s_1$ divided by query character 3-gram count.
\end{enumerate}
The logistic model adds one intercept, for eight parameters total. Five-fold
splits hold out skills, while meta examples split by paired utterance ID.

\section{User-Response Model}
\label{app:user}

The frozen simulation uses seed 0 and one top-2 retry.
\begin{enumerate}
  \item The confirmation path presents top-1, then at most one top-2 candidate.
  \item The correct button is yes exactly when the candidate equals gold and
  gold is a skill or file tool.
  \item Chat, ask, and delegate gold labels require no.
  \item The user presses the correct button with probability $u$ and the
  opposite button otherwise, for $u\in\{1.0,0.9,0.8\}$.
  \item A yes with missing required slots becomes \code{meta\_ask}.
  Otherwise it executes the candidate.
  \item A final no becomes \code{meta\_delegate}.
  \item Below $t_{\mathrm{low}}$, a missing slot also becomes
  \code{meta\_ask} without another model call.
\end{enumerate}

\section{Deployment Errors}
\label{app:errors}

\begin{samepage}
\paragraph{Gemma 4 E4B-it.}
The captured terminal exception follows.
\begin{quote}\scriptsize\ttfamily\raggedright
AmbiguousGlobal\allowbreak PerLayer\allowbreak AttributeError:\\
'head\_dim' is a per-layer attribute and may vary across layers.
Access it via the individual layer configs instead
(e.g. config.per\_layer\_config[i].\allowbreak head\_dim).
\end{quote}
The stack reaches \code{vLLM}'s model architecture converter, calls
\code{get\_head\_size}, and reads the global \code{head\_dim}. The load fails
before weight admission.
\end{samepage}

\paragraph{Granite 4.0 H-Small.}
The runtime parser registry contains \code{granite} and \code{granite4}.
It does not contain \code{granite\_pythonic}, which the checkpoint's
Python-style calls require. The XML-oriented \code{granite4} parser is not a
substitute. The BF16 checkpoint also exceeds 23,034 MiB. We do not run a
quantized replacement.

\section{Complete Condition Tables}
\label{app:alltables}

The remaining pipeline label, P4, uses BGE for both ranking and gating. It
reaches the highest completion in the table, 478 of 600 against P1's 453, and it
also delegates 26 more requests, 250 against 224. Of its 25 extra completions,
20 are delegated local tasks, which completion credits, and five are exact
resolutions. Its coverage is 350 of 600, twenty-six requests below P1, so the
adoption rule keeps P1, and its danger rate is 8 of 150 delegates rather than
the 9 of 150 the other four hold.
Table~\ref{tab:pipelines} reports all five pipelines at the product operating
point, Table~\ref{tab:candidates-language} splits the candidate ablation by
language, and Table~\ref{tab:native-conditions} lists both native tool-calling
conditions.

\begin{table*}[t]
\centering
\caption{P3 exceeds P1 by nine requests on accuracy and eight on completion at
the same danger and ask rates, and by a single request on coverage, 377 of 600
against 376. Section~\ref{sec:rerank} rejects it on measured cost rather than on
a metric. P4 leads the completion column by delegating 26 more requests, which
completion credits. The text above decomposes that lead. Results use $u=1$, ask
at most 0.200, and danger at most 0.060.}
\label{tab:pipelines}
\small
\begin{tabular}{l S{1.3} S{1.3} S{1.3} S{1.3} S{1.3} S{3.1} S{3.1} S{1.0}}
\toprule
ID & \multicolumn{1}{c}{Accuracy} & \multicolumn{1}{c}{Completion} &
\multicolumn{1}{c}{Coverage} & \multicolumn{1}{c}{Danger} &
\multicolumn{1}{c}{Ask} & \multicolumn{1}{c}{CPU (ms)} &
\multicolumn{1}{c}{MB} & \multicolumn{1}{c}{Models} \\
\midrule
P1 & 0.645 & 0.755 & 0.627 & 0.060 & 0.198 & 13.9 & 278.3 & 1 \\
P2 & 0.637 & 0.767 & 0.607 & 0.060 & 0.172 & 13.7 & 278.3 & 1 \\
P3 & 0.660 & 0.768 & 0.628 & 0.060 & 0.198 & 197.3 & 846.1 & 2 \\
P4 & 0.653 & 0.797 & 0.583 & 0.053 & 0.185 & 183.6 & 567.8 & 1 \\
P5 & 0.642 & 0.763 & 0.615 & 0.060 & 0.172 & 13.9 & 278.3 & 1 \\
\bottomrule
\end{tabular}
\end{table*}

\begin{table*}[t]
\centering
\caption{Candidate restriction improves mismatch selection in both languages.
Values are means and sample standard deviations over five seeds. Only the
paraphrase slice is split by language. The pooled matched slice stays at or above
0.951 at every candidate count in Table~\ref{tab:candidates}.}
\label{tab:candidates-language}
\small
\begin{tabular}{S{2.0} *{8}{S{1.3}}}
\toprule
\multicolumn{1}{c}{$k$} &
\multicolumn{2}{c}{BM25 KO} & \multicolumn{2}{c}{BM25 EN} &
\multicolumn{2}{c}{e5 KO} & \multicolumn{2}{c}{e5 EN} \\
\cmidrule(lr){2-3}\cmidrule(lr){4-5}\cmidrule(lr){6-7}\cmidrule(lr){8-9}
& \multicolumn{1}{c}{Mean} & \multicolumn{1}{c}{SD} &
\multicolumn{1}{c}{Mean} & \multicolumn{1}{c}{SD} &
\multicolumn{1}{c}{Mean} & \multicolumn{1}{c}{SD} &
\multicolumn{1}{c}{Mean} & \multicolumn{1}{c}{SD} \\
\midrule
7 & 0.814 & 0.029 & 0.836 & 0.038 & 0.814 & 0.011 & 0.855 & 0.028 \\
10 & 0.798 & 0.027 & 0.795 & 0.033 & 0.759 & 0.031 & 0.822 & 0.037 \\
20 & 0.667 & 0.023 & 0.687 & 0.028 & 0.699 & 0.024 & 0.749 & 0.036 \\
40 & 0.581 & 0.038 & 0.624 & 0.031 & 0.651 & 0.023 & 0.684 & 0.018 \\
70 & 0.482 & 0.000 & 0.542 & 0.000 & 0.590 & 0.000 & 0.627 & 0.000 \\
\bottomrule
\end{tabular}
\end{table*}

\begin{table*}[t]
\centering
\caption{Top-5 retrieval improves LFM2.5 routing and reduces its no-call rate,
while Qwen changes little. All and No call are over the 600 requests including
meta labels; Match and Mismatch are over the 164 and 166 rankable requests, so
the three accuracy columns do not form a weighted average.}
\label{tab:native-conditions}
\small
\begin{tabular}{ll S{1.3} S{1.3} S{1.3} S{1.3} S{4.0}}
\toprule
Model & Tools & \multicolumn{1}{c}{All} & \multicolumn{1}{c}{Match} &
\multicolumn{1}{c}{Mismatch} & \multicolumn{1}{c}{No call} &
\multicolumn{1}{c}{GPU (ms)} \\
\midrule
LFM2.5-2.6B & full & 0.442 & 0.774 & 0.530 & 0.143 & 882 \\
LFM2.5-2.6B & top-5 & 0.540 & 0.915 & 0.753 & 0.098 & 984 \\
Qwen3-4B & full & 0.673 & 0.976 & 0.801 & 0.000 &
\multicolumn{1}{r}{1,152} \\
Qwen3-4B & top-5 & 0.705 & 0.957 & 0.777 & 0.000 &
\multicolumn{1}{r}{1,111} \\
\bottomrule
\end{tabular}
\end{table*}

\end{document}